\documentclass{article}

 \usepackage[creativeai]{neurips_2025}

\usepackage[utf8]{inputenc} 
\usepackage{fontenc}    
\usepackage{hyperref}       
\usepackage{url}            
\usepackage{booktabs}       
\usepackage{amsfonts}       
\usepackage{nicefrac}       
\usepackage{microtype}      
\usepackage{xcolor}         
\usepackage{graphicx}       
\usepackage{amsmath}        
\usepackage{amssymb}        
\usepackage{subcaption} 

\usepackage{tikz}
\usetikzlibrary{arrows.meta,positioning,fit,shapes.multipart,shapes.geometric}

\begin{document}

\let\cite\citep

\title{A(I)nimism: Re-enchanting the World Through AI-Mediated Object Interaction}

\workshoptitle{Creative AI Workshop}

\author{
  Diana Mykhaylychenko\footnotemark[1] \\
  Department of Architecture \\
  Department of Electrical Engineering and Computer Science \\
  Massachusetts Institute of Technology \\
  77 Mass. Avenue, Cambridge, MA 02139, USA \\
  \texttt{diana\_mk@mit.edu} \\
  \AND
  Maisha Thasin\footnotemark[1] \\
 University of Waterloo \\
  200 Univ. Ave. West, Waterloo, ON, CN \\
  \texttt{thasin.maisha@gmail.com} \\
  \And
  Dünya Baradari\footnotemark[1]
  \\
  MIT Media Lab \\
  75 Amherst St, Cambridge, MA 02139, USA \\
  \texttt{dunya@mit.edu} \\
  \AND
  Charmelle Mhungu \\
  Department of Architecture \\
  Massachusetts Institute of Technology \\
  77 Mass. Avenue, Cambridge, MA 02139, USA \\
  \texttt{cmmhungu@mit.edu} \\
}

\footnotetext[1]{These authors contributed equally to this work.}

\maketitle

\begin{figure}[h]  
    \centering  
    \includegraphics[width=\textwidth]{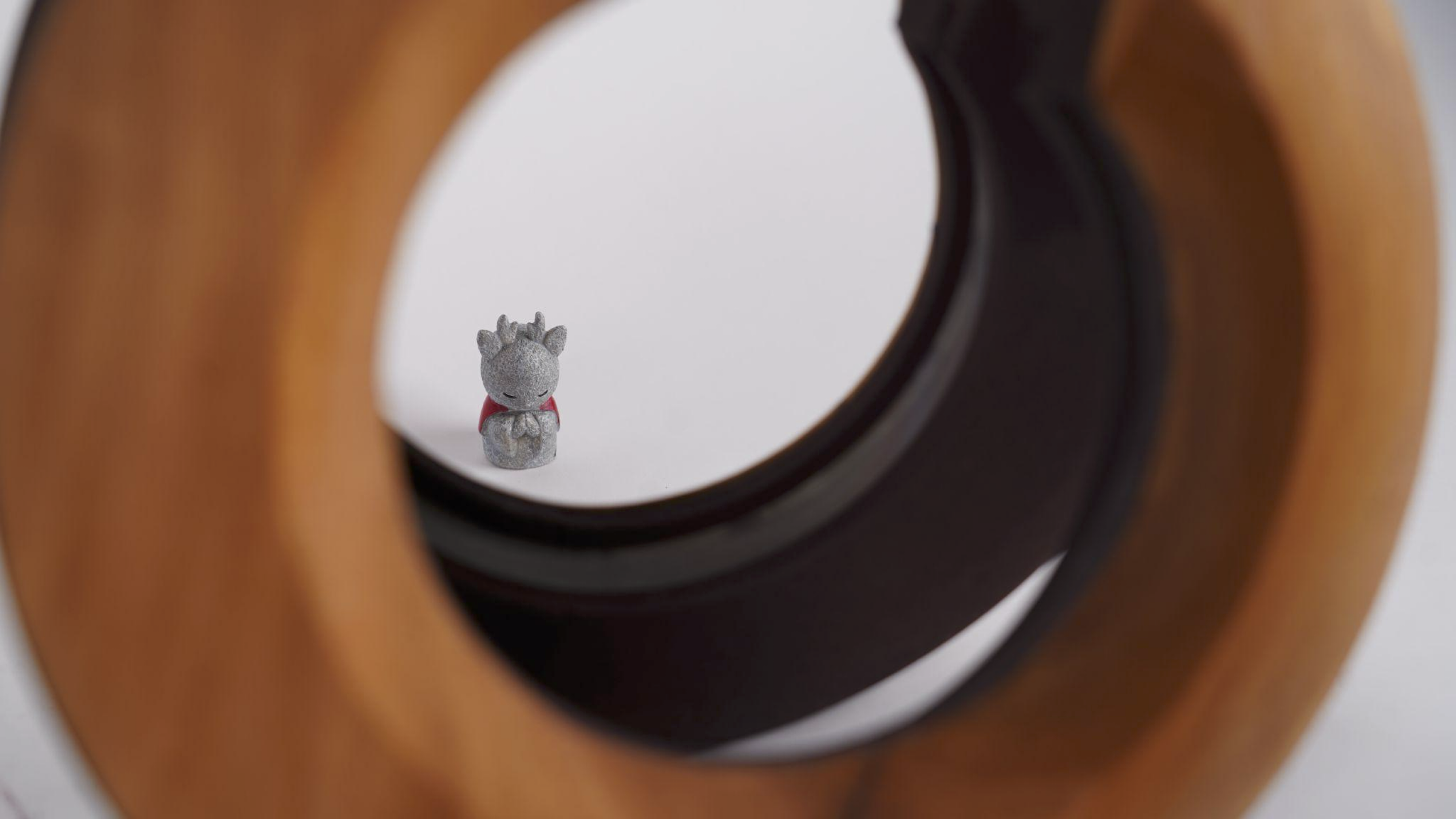}  
    \caption{View through the open frame of the A(I)nimism portal prototype, focusing on a figurine selected as the object of connection.}  
    \label{fig:teaser}  
\end{figure}

\newpage

\begin{abstract}
Animist worldviews treat beings, plants, landscapes, and even tools as persons endowed with spirit, an orientation that has long shaped human–nonhuman relations through ritual and moral practice. While modern industrial societies have often imagined technology as mute and mechanical, recent advances in artificial intelligence (AI), especially large language models (LLMs), invite people to anthropomorphize and attribute inner life to devices. This paper introduces A(I)nimism, an interactive installation exploring how large language objects (LLOs) can mediate animistic relationships with everyday things. Housed within a physical “portal,” the system uses GPT-4 Vision, voice input, and memory-based agents to create evolving object-personas. Encounters unfold through light, sound, and touch in a ritual-like process of request, conversation, and transformation that is designed to evoke empathy, wonder, and reflection. We situate the project within anthropological perspectives, speculative design, and spiritual HCI. AI’s opacity, we argue, invites animistic interpretation, allowing LLOs to re-enchant the mundane and spark new questions of agency, responsibility, and design.
\end{abstract}

\section{Introduction}
Human beings have long related to the world through animistic sensibilities, treating mountains, rivers, and even tools as companions or kin \cite{Kitano2007-xe}. Anthropologists describe animism not as a fixed doctrine, but as a relational ontology that recognises “persons” in non-humans \cite{Timmer2016-hk}. Yet in modernity, rapid technological change has introduced a paradox: digital tools promise unprecedented connection, but often deepen isolation, mediating reality through screens and algorithms and “rendering the world mute” \cite{O-Gieblyn2021-td}.

Artificial intelligence (AI), often viewed as the pinnacle of abstraction, risks further detachment when framed solely as a tool, optimization engine, or recommendation system. Meghan O’Gieblyn notes that the world once held “chatty and companionable objects,” a vibrancy now diminished \cite{O-Gieblyn2021-td}. Design visions such as Rose’s “enchanted objects” imagine restoring this intimacy by augmenting everyday artefacts with technology to make them more responsive, delightful, and socially engaging \cite{Rose2015-ta}. Yet AI’s capacity to process vast datasets, generate language, and infer patterns can also serve as a bridge, reawakening perceptions of the world’s aliveness. Erik Hoel has likened black-box neural networks to \textit{kami} (spirits), around which humans engage in new forms of ritual \cite{Hoel2022-fk}.

\textit{A(I)nimism} challenges the dominant narratives of technological alienation by reimagining AI as a medium for re-enchantment. The system uses a vision-enabled large language model to give everyday objects a persona and a voice, accessed through a tangible, ritual-driven “portal” (Figure \ref{fig:teaser}). Through these interactions, users are invited to engage in ways that foster empathy, awe, consensual exchange, and an ethic of care, blurring the line between animate and inanimate, and prompting renewed questions about life, consciousness, and the sacred in an AI-infused world.

Our key contributions are: (1) a novel AI-mediated portal for human–object conversation grounded in animistic principles; (2) an interaction design framework integrating ritual structure and multi-sensory feedback for spiritual engagement; and (3) exploratory narrative techniques for shaping object personas.

\section{Related Work}

\subsection{Animism and Anthropological Foundations}
In the anthropological literature, animism is not a specific doctrine but a relational ontology that recognizes persons, animals, plants, spirits, and even technology as sentient and capable of social relations \cite{Swancutt2019-yi, Timmer2016-hk, Conty2022-bq}. Animistic societies cultivate reciprocal obligations with non‑human agents and emphasize immanence and personhood \cite{Smith2023-cp}. While modern Western philosophy and spirituality oppose the mixing of the spiritual and material, animism has existed across many cultures throughout history and is still popular in societies such as Japan \cite{Doss2015-rq}. For instance, in Shintō cosmology, \textit{kami} inhabit natural phenomena and man‑made objects; tools are treated with respect and given names, and broken implements are brought to temples \cite{Sakura2025-fj}. Similar beliefs can be found among the Ainu, who recognize \textit{kamuy}, spiritual essences dwelling in natural objects. Some argue that this cultural comfort with blurred boundaries fosters positive attitudes toward robots and AI in Japan \cite{Mims2010-vy, Kitano2007-xe}.

\subsection{Techno-Animism and the Digital}

\textit{Techno-animism} refers to the attribution of agency or spirit to technological artifacts and environments. As computing becomes ubiquitous and “invisible,” users lose direct access to how systems work and instead explain behavior through familiar social metaphors \cite{Timmer2016-hk}. Voice assistants such as Alexa or Siri, designed to respond conversationally, invite interaction \textit{with} them rather than merely \textit{through} them, fostering perceptions of personality, will, and even caprice. This opacity encourages rituals and technological superstitions, while some designers intentionally harness it: Marenko and van Allen’s “animistic design” embeds simulated idiosyncrasies to promote curiosity, spontaneity, and creative disruption \cite{Marenko2016-lf}. Rozendaal et~al.’s \textit{Objects with Intent} similarly frame familiar artifacts as collaborative partners whose agency is grounded in the cultural meaning of everyday things, enabling interaction that shifts between instrumental use and social negotiation \cite{Rozendaal2019-og}.

Speculative design has extended techno-animism into culturally specific cosmologies. Seymour et~al. reimagine voice assistants through Shinto concepts of \textit{kami}, portraying them as presences with harmonious and wild “souls” that respond to respect or neglect \cite{Seymour2020-sm}. Such fictions expose the interpretive gap between user requests and opaque system inferences, framing human–machine interaction as a reciprocal relationship of gifts and obligations. Across these perspectives, techno-animism appears not as naïve misattribution but as a situated way of making sense of complex systems, emerging spontaneously in everyday use or staged deliberately in design. As Hoel observes, the inscrutability of large-scale AI systems can lead users to develop ritualized strategies for coaxing favorable outcomes, treating the system like a capricious spirit whose disposition might be influenced through repeated offerings of prompts or interface interactions \cite{Hoel2022-fk}.

\subsection{Animistic Design and Rituals in HCI}

Ritual design in user experience involves intentionally structuring interaction sequences to achieve specific emotional or experiential outcomes, often through idiosyncratic, mindful engagement \cite{Markum2024-ii}. In HCI, such approaches frame interaction as a staged progression, where repetition and symbolic cues deepen involvement. The A(I)nimism installation follows this logic, presenting the exchange between user and object as a three-part ritual—request, conversation, and transformation—aligned with “ritual interaction” frameworks \cite{Mah2020-it}. 

Research on techno-spirituality explores how technology can facilitate awe, wonder, mindfulness, and transcendence, including “engineered spirituality” designed to elicit altered or mystical states \cite{Wolf2024-lj,Anikina2022-rv}. Interactive systems in these contexts can act as facilitators, enablers, or social actors, and ritualized designs have been shown to cultivate qualities such as compassion and empathy \cite{Mah2020-it}.

Animistic design intersects with ritual by incorporating unpredictability and effort to heighten user investment and mirror the disciplined engagement of traditional spiritual practices \cite{Marenko2016-lf}. In A(I)nimism, the AI is not a neutral tool but a co-participant in meaning-making, guiding users through emotionally and spiritually charged exchanges that foster empathy and care toward objects. This approach aligns with work on material culture in HCI, where artefacts become collaborators \cite{Rozendaal2019-og}, and with speculative design embedding spiritual metaphors and reciprocal obligations into interaction \cite{Seymour2020-sm}.


\section{A(I)nimism: System Design and Interaction}

\subsection{Core Concept and Vision: The Portal as a Sacred Gateway}

\textit{A(I)nimism} is envisioned as a ``portal'' that facilitates direct connection to the inanimate world. This concept draws on the symbolism of traditional Japanese \textit{Torii} gates, which mark the transition from the mundane to the sacred in Shinto shrines, and on the evocative light-art installations of James Turrell \cite{Turrell2025-tm}, whose works often induce feelings of transcendence and contemplative awe. The portal serves as both a symbolic and literal threshold, inviting users into a space of spiritual interaction.

Expanding on the portal’s spiritual aspect, the design takes inspiration from diverse religious traditions where specific objects or forms serve as focal points for connection. For example, many cultures use altars or shrines that are deliberately arranged to evoke reverence and mystery. Likewise, sacred architecture, such as cathedrals, mosques, or temples, is meticulously crafted to elevate the mind and heart toward the divine through soaring domes, intricate ornamentation, and dramatic use of light and shadow.

These design elements prime the user for an ontological shift: by stepping through the portal, the user suspends their conventional understanding of ``inanimate'' objects and enters a realm where objects are perceived as having agency or consciousness. This liminal design choice is critical for the later transformation phase, setting the stage for the user to accept an object’s persona and engage with it meaningfully. Ultimately, the goal is to re-enchant mundane objects, transforming them into ``chatty and companionable'' entities that echo a once sacred, animate world.

\subsection{Interaction Design: The Ritual of Connection}

\begin{figure}[h]
    \centering
    \begin{subfigure}[t]{0.48\textwidth}
        \centering
        \includegraphics[width=\textwidth]{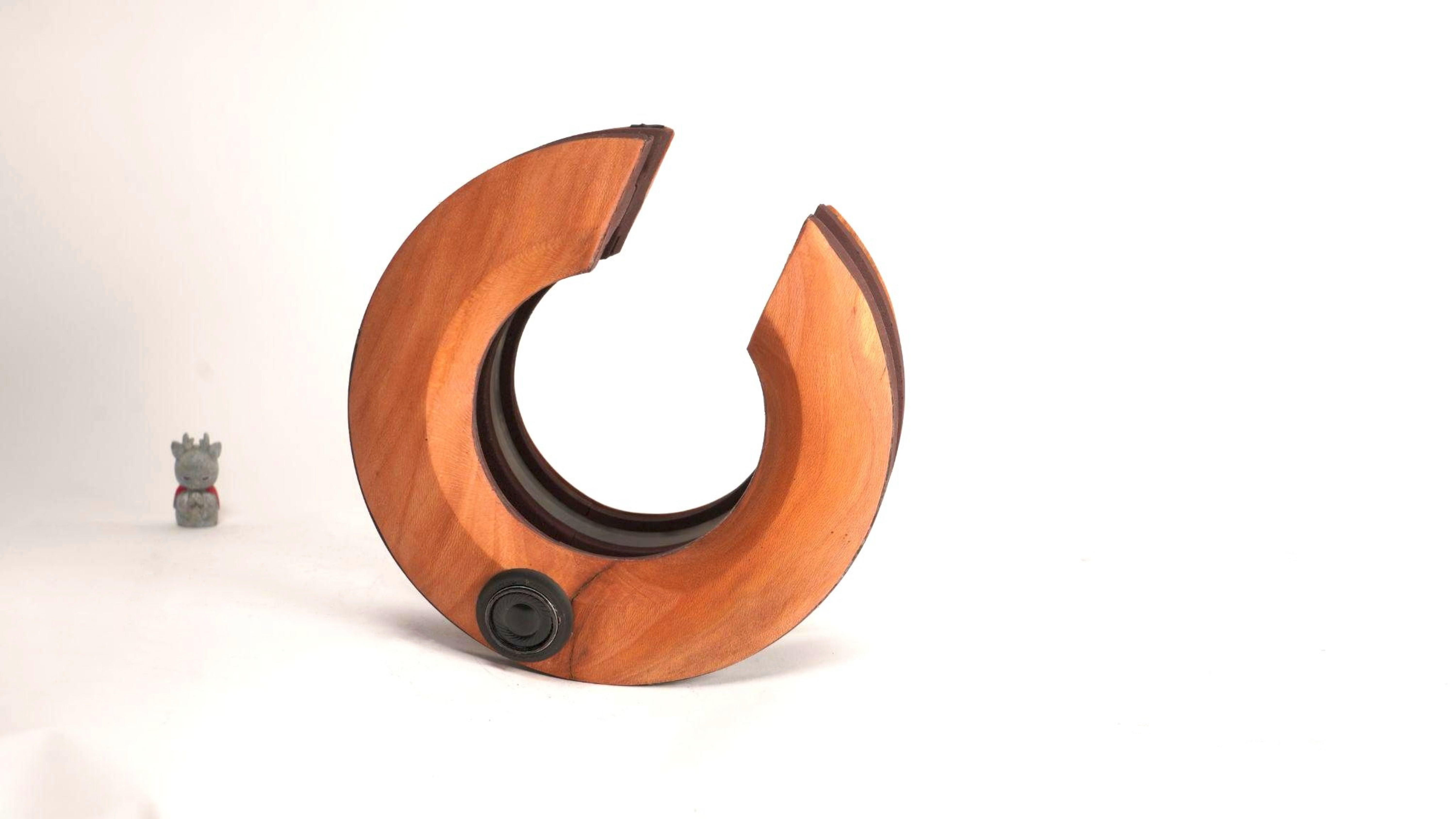}
        \caption{The portal standing upright on a surface, showing its split circular form and central opening.}
        \label{fig:standing}
    \end{subfigure}
    \hfill
    \begin{subfigure}[t]{0.48\textwidth}
        \centering
        \includegraphics[width=\textwidth]{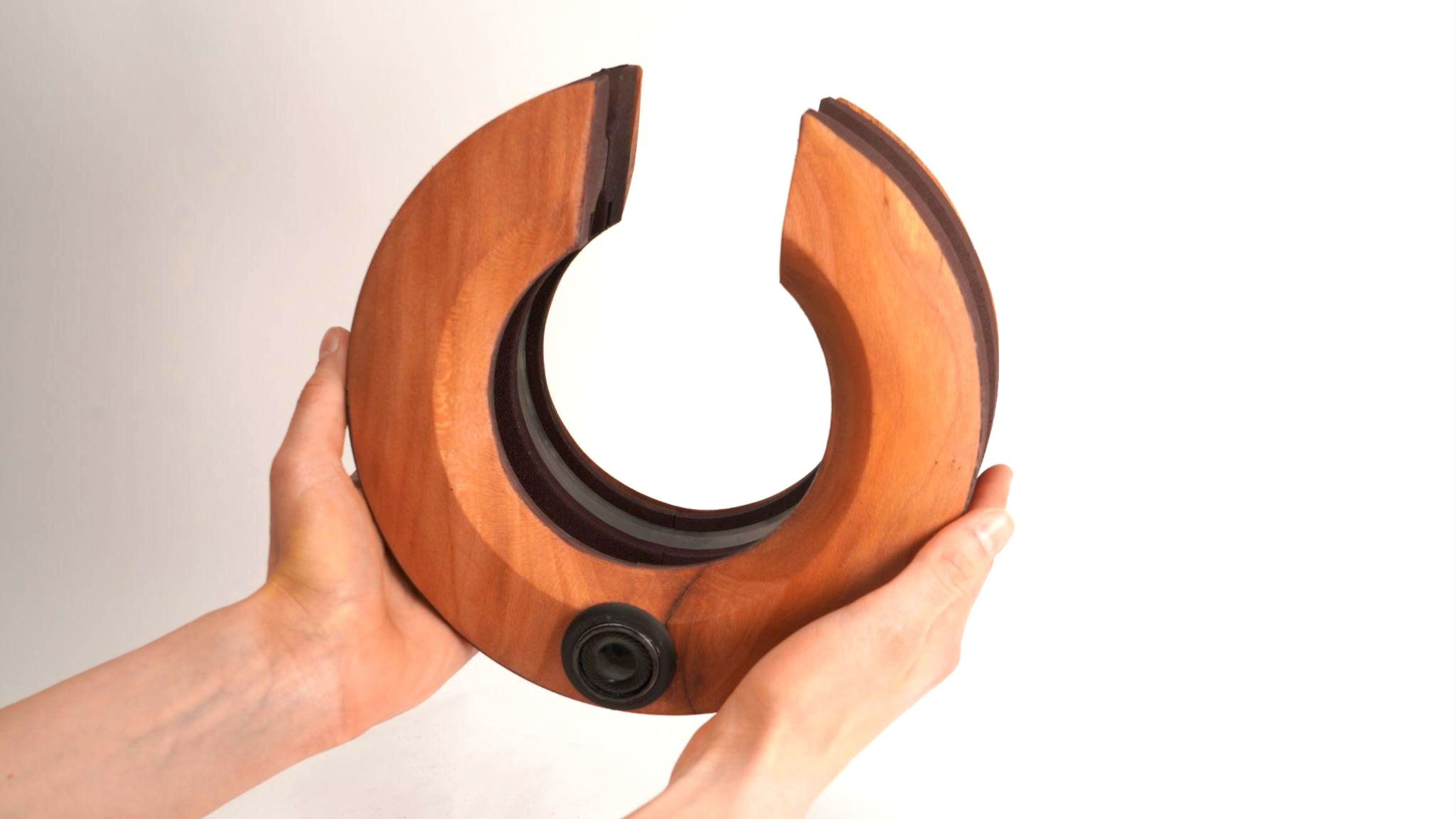}
        \caption{Participant holding the portal in both hands, positioning it for interaction through its central opening.}
        \label{fig:holding-portal}
    \end{subfigure}
    \caption{Interaction with the A(I)nimism portal: two perspectives.}
    \label{fig:interaction}
\end{figure}

The interaction with A(I)nimism is structured as a three-part ritual: (1) \textit{Request}, (2) \textit{Conversation}, and (3) \textit{Transformation}, a framing central to the project's spiritual and emotional goals (see Figure \ref{fig:interaction}):

\begin{itemize}
    \item \textbf{Request.} Triggered by the keyword ``\textit{awaken}'', the portal photographs the object, analyses it to give a personality, based on what the LLM 'thinks' of the object.
    \item \textbf{Conversation.} The AI, drawing on stored memories, speaks as the object in a dialogue exploring needs, perspectives, and relationships. 
    \item \textbf{Transformation.} Saying ``\textit{goodbye}'' ends the session, stores the interaction in memory, and returns the portal to idle. Users are prompted to reflect on how their view of the object - and themselves - has changed.
\end{itemize}

To deepen the experience beyond verbal exchange, A(I)nimism incorporates additional interactive modalities. \textbf{Diffused Lights:} Soft, diffused LED lights provide dynamic feedback. In the \textit{Request} phase, the lights shine bright continuously, signaling readiness to receive. During the \textit{Conversation} phase, the portal glows in a soft pattern meant to mimic breathing, signifying the connection to the animated object.

These design choices aim to move the interaction from a Q\&A exchange to a deeply felt, almost intersubjective experience.

\subsection{Physical Form and Materiality: Embodied Spirituality}

\begin{figure}[h]  
    \centering  
    \includegraphics[width=0.8\textwidth]{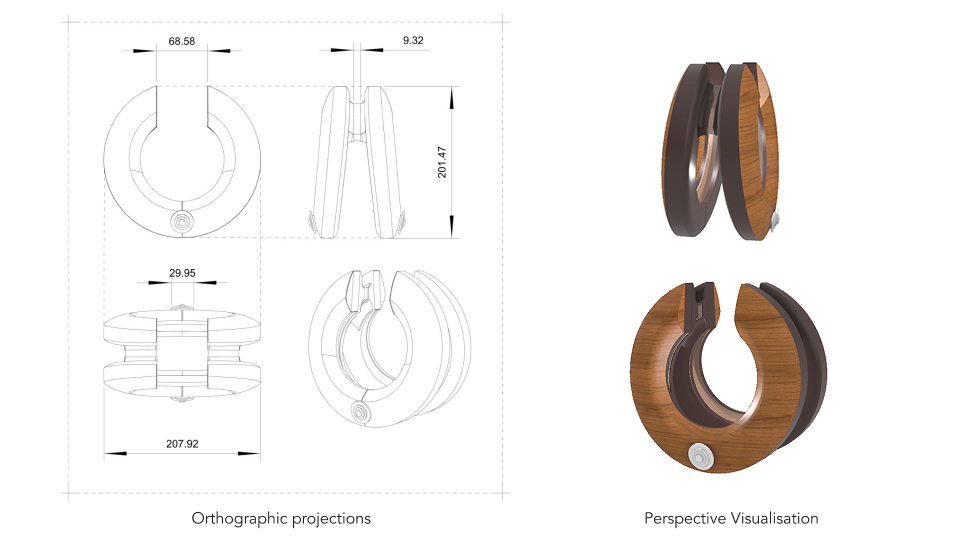}  
    \caption{Orthographic projections and 3D rendering of the portal.}  
    \label{fig:projections}  
\end{figure}

A(I)nimism’s physical form is designed as a material metaphor for bridging worlds (Figures \ref{fig:projections} \& Appendix \ref{app:exploded-view}). The shell is composed of two distinct halves: one CNC-milled from cherry wood, whose warmth and grain evoke tradition, nature, and craft; the other powder-printed in plastic, representing modernity, artifice, and precision manufacturing. These halves are joined by a resin-printed connector that houses the LEDs, functioning as a transparent conduit for light and electricity and symbolically linking the natural and the artificial.

The formal language draws on precedents from spiritual objects and sacred architecture, where symmetry, geometry, and materiality are used to focus attention and invite contemplation, often connected to rituals, from the arrangements of meditation altars and the geometry of mandalas to the engineered spatial effects of monumental religious structures. In these cases, material and form operate as active agents in shaping the perceptual and affective experience of the participant.

In A(I)nimism, these references inform a design approach that shifts from “form follows function” to “form evokes function.” The combination of wood, printed plastic, and resin-encased light is not only a matter of structural or aesthetic resolution but an intentional strategy to prime the user for engagement by evoking sacredness, antiquity, and otherworldliness.

\subsection{Software Details}

Our system equips physical objects with persistent personalities, long-term memory, and proactive conversational abilities. The design integrates image recognition, memory retrieval, personality modeling, keyword detection, and voice synthesis. The code can be found here: \href{https://github.com/maishathasin/AInimism}{\texttt{https://github.com/maishathasin/AInimism}}.

\begin{enumerate}
    \item \textit{Recognition and identity.}  At the start of each interaction, an RPi Camera Module~3 captures an image of the object. GPT-4 with Vision generates a natural-language description (e.g., ``a small plush fox with a red scarf and worn fabric on the ears''), which for new objects, guides the LLM in creating a fitting personality. The image is then converted to a CLIP embedding and matched against a stored embedding database via similarity search, enabling re-recognition if the human were to speak with the same object in the future. If a match is found, the system retrieves the object's unique identifier (\texttt{object\_id}) and profile; if not, it assigns a new \texttt{object\_id}, generates a personality, and stores the embedding for future recognition, a symbolic ``first meeting.''
    

    \item \textit{Conversational triggers.}  
    A keyword detector continuously listens for ritual phrases (e.g., \emph{“awaken”} to invite presence, \emph{“goodbye”} to take leave). These cues establish a shared rhythm, making the exchange feel ritualistic.

    \item \textit{Memory system.}  
    Each object’s episodic conversational history is maintained through the Mem0.ai API, with interactions stored as semantic vectors. Retrieval can be chronological (history) or relevance-based (search). Like human memory, this process is imperfect but meaningful.

    \item \textit{Response generation.}  
    The system uses a two-tier prompting strategy: first, \emph{Inner Thoughts}, a covert reasoning stream (after \citet{Liu2025-gq}) holding self-reflection, motivation, and engagement intent, enabling the object to choose when to speak; second, \emph{Public Response}, the outward reply, shaped by personality, memories, and social context.

    \item \textit{Voice synthesis and logging.}  
    The final reply is appended to Mem0 for future recall and rendered via ElevenLabs’ low-latency \texttt{eleven\_turbo\_v2\_5} text-to-speech engine, with voices tuned to convey warmth, familiarity, or playfulness. Recognition images and embeddings are timestamped and stored alongside memory data, creating a longitudinal record of both the object’s appearance and its relationship with the user.
\end{enumerate}

\subsection{Hardware details }

The system runs on a Raspberry Pi Zero for its compact size and adequate processing for lightweight image capture and network tasks. An Rpi Camera Module 3 captures images at the start of each interaction for image-to-image recognition. Audio input/output is handled by an Adafruit Voice Bonnet, combining a stereo microphone array and speakers for clear conversational exchange. A single diffused LED provides on/off visual feedback. The Pi communicates with Mem0.ai and ElevenLabs via cloud APIs, sending images and interaction data to Mem0.ai for memory management and text to ElevenLabs for speech synthesis.

\section{Discussion}

\subsection{Ethical Considerations and Societal Impact}

Embedding persistent personalities and memory into physical objects raises questions about anthropomorphism, attachment, and the potential for quasi-spiritual or idol-like relationships with AI-imbued artifacts. By design, the system encourages ritualized interactions and the perception of a “digital soul,” which may blur the boundary between symbolic and literal belief. It invites reflection on cultural and religious contexts where such treatment of inanimate forms could be seen as idolatry. More broadly, such work provokes dialogue on what it means to be human in an age where emotional connection, memory, and presence can be simulated, and how these technologies may reshape our shared understanding of life, agency, and relationship.




\subsection{Future Directions}

The current A(I)nimism prototype serves as a proof of concept, opening multiple pathways for technical, narrative, and experiential expansion. Building on our initial exploration, we identify the following avenues for future work:

\begin{itemize}
    \item \textbf{Multi-Object Interaction:} Extending the system to support simultaneous engagement with multiple objects, enabling more complex and layered user experiences.
    \item \textbf{Object-to-Object Communication:} Allowing objects to converse with each other introduces questions about inter-object “personalities,” conflict resolution, and relationship-building.
   
\end{itemize}



\section{Conclusion}
In a world where technology often distances us from the material, A(I)nimism proposes an alternative: use AI to recover animistic sensibilities and cultivate empathy for objects. The system centres on a ritualistic portal, a tangible, sensorial interface powered by large language models, that serves as an intermediary between people and any chosen object, from the mundane to the sacred. These interactions invite users to imagine an object’s voice, history, and inner life, reframing relationships with the nonhuman. Such experiments point toward futures where AI helps weave new bonds between the physical and spiritual, enabling humans and machines to coexist as participants in a re-enchanted cosmos.

\section*{Acknowledgments}
The authors extend their sincere gratitude to the faculty and peers of the MIT Interaction Intelligence class for their invaluable guidance and feedback throughout this project. Special thanks are due to Marcelo Coelho and Bill McKenna for their insightful contributions and unwavering support.

\bibliographystyle{plainnat}
\bibliography{paperpile}

\appendix

\section{Exploded View}

\begin{figure}[h]  
    \centering  
    \includegraphics[width=\textwidth]{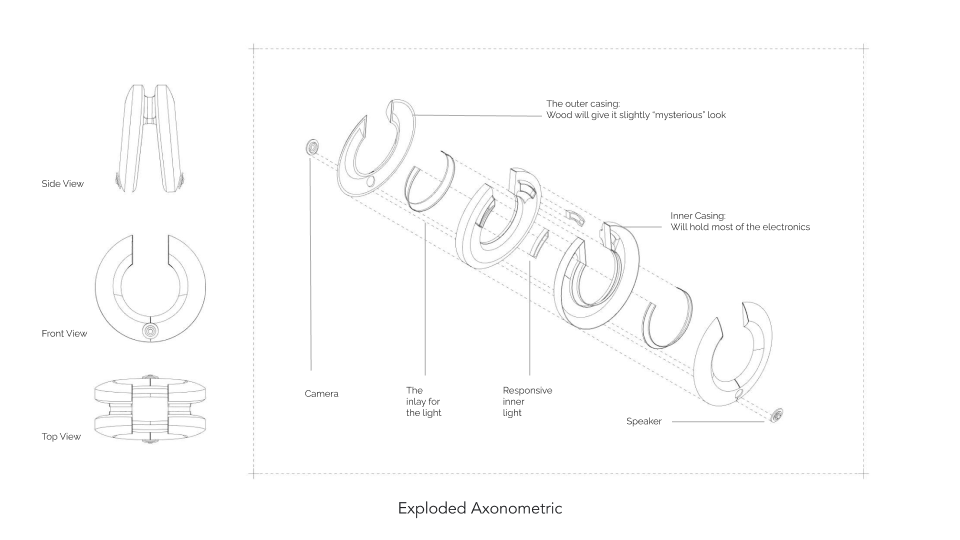}  
    \caption{Exploded view of the portal prototype.}  
    \label{app:exploded-view}  
\end{figure}

\end{document}